# COMPARISON OF TURKISH WORD REPRESENTATIONS TRAINED ON DIFFERENT MORPHOLOGICAL FORMS

Gökhan Güler and A. Cüneyd Tantuğ

Department of Computer Engineering, Istanbul Technical University,
Istanbul, Turkey

## ABSTRACT

*Increased popularity of different text representations has also brought many improvements in Natural Language Processing (NLP) tasks. Without need of supervised data, embeddings trained on large corpora provide us meaningful relations to be used on different NLP tasks. Even though training these vectors is relatively easy with recent methods, information gained from the data heavily depends on the structure of the corpus language. Since the popularly researched languages have a similar morphological structure, problems occurring for morphologically rich languages are mainly disregarded in studies. For morphologically rich languages, context-free word vectors ignore morphological structure of languages. In this study, we prepared texts in morphologically different forms in a morphologically rich language, Turkish, and compared the results on different intrinsic and extrinsic tasks. To see the effect of morphological structure, we trained word2vec model on texts which lemma and suffixes are treated differently. We also trained subword model fastText and compared the embeddings on word analogy, text classification, sentimental analysis, and language model tasks.*

## 1. INTRODUCTION

Text representations are essential to use texts in computations for different NLP tasks and come up in many different representation methods. One earlier popular solution was one-hot representation: vectors with size of the vocabulary that only vocabulary index of the corresponding word is set as 1 while all other elements are 0. However, bloated size of vectors and not being able to preserve any relational meaning between vectors led researchers to develop another solution: continuous word embeddings.

Continuous word embeddings are vectors with variable length and contain real numbers instead of zeros and ones. It is possible with continuous word vectors to contain meaningful relations between vectors. Obtaining these vectors is generally done by training by different kinds of models. One problem with these models are that they treat each token as a different word, disregarding they might be the same word with different word forms. Each word form is treated as a different word and has a different vector. This can be questioned as dividing the meaning of a word into different forms. Another issue is that with the increased number of word forms, occurrences for each token decrease in parallel. This is not a big problem for English or other European languages since they have relatively limited vocabulary size. However, for agglutinative languages like Turkish, Finnish and Hungarian, more than a thousand word forms can be generated from just a single lemma. Due to very productive inflectional and derivational morphology, the size of the corpus should be infeasibly large to cover all word forms. Even in a large corpus, this morphological productivity leads large number of out-of-vocabulary words.

In this study, we have trained context-free word embeddings on different morphological forms of words and compared the results with the basic continuous word embeddings and contextual embeddings on tasks of word analogy, text classification, sentimental analysis and language modelling. In Section 2, we have listed the studies related to our experiments. In Section 3, we explained the embedding models we have used to train and in Section 4, we explained what kind of representation models we have derived from them. We discussed the test result in Section 5 and the conclusion and possible future works in Section 6.

## 2. RELATED WORK

Rumelhart et al. [1] offered one of the first solutions to context-free word representations. Bengio et al. [2] also used this architecture successfully. Remarkably, Mikolov et al. [3] proposed simple yet powerful neural network architecture called Skip-Gram and CBOW models to train continuous word representations. Words with their contexts are given to a neural network with one hidden layer to train word embeddings, in other words, word vectors. Le and Mikolov [4] used the same models for texts to get vectors of text blocks and Ling et al. [5] proposed Structural Skip-Gram/CBOW models where context vectors are concatenated instead of being averaged. Pennington et al. [6] used global co-occurrences to train word vectors in their GloVe model.

These models have also been studied for Turkish language as well. Sen and Erdogan [7] has tested word2vec model [3] with various settings on a Turkish corpus and shared the results of each setting change. Güngör and Yıldız [8] have examined the same model on a more detailed test sets and showed the results for each kind of word derivation separately. Erdinç and Güran [9] have used word2vec for text classification by different classifier models and compared the results. Aydoğan and Karcı [10] compared word2vec [3] and GloVe [6] models on a large analogy task test set.

Morphological structures have also been included in word embedding studies. Luong et al. [11] used a model where input is morphologically segmented and given to a Recursive Neural Network. Instead of training a vector for each word, they trained a vector for each morpheme. Similarly, Botha and Blunsom [12] used morphemes in a Log-Bilinear Language Model to train vectors. Ling et al. [13] character-based word embeddings and trained them in bi-directional LSTMs. Bojanowski et al. [14] trained subword vectors in CBOW model [3] to obtain word embeddings.

## 3. REPRESENTATION MODELS

### 3.1. word2vec

First proposed by Mikolov et al. [3] and developed with hierarchical softmax, negative sampling, subsampling of frequent words and learning phrases in [15]. This model can produce word vectors with meaningful relations. One example for this from the paper is that difference between the vectors for "man" and "woman" successfully preserves the difference between manhood and womanhood so that if we add this difference vector to the vector of "king", we obtain a vector very close to the vector of "queen" compared to any other word vector.

word2vec model proposes a simple neural network with one hidden layer with an arbitrary number of nodes. This model has two variations: Continuous Bag-of-Words (CBOW) and Skip-Gram. In training, data is iterated over with a sliding window. The word in the middle of window is called target word and other words are called context words. At each iteration, target and context words are given as input and output to the neural network in one-hot representation.

In CBOW, context words are given as input and the output is expected to get close to target word's one-hot vector. In other words, output becomes a vector of possibilities for each word in vocabulary where probability of the target word is close to 1 and all other words' probabilities are close to 0. Softmax function is used to fit output layer values between 0 and 1. In Skip-Gram, target word is given as input and output is expected to get close to context words' one-hot vectors.

With a window size of 5 and vocabulary size of V, CBOW has an input layer with 4 x V nodes and output layer with V nodes while Skip-Gram has an input layer with V nodes and output layer with 4 x V nodes. After training with back propagation algorithm, the first layer's parameter matrix gives us the word vectors, each row being a word vector. The second layer's parameter matrix also provides us a set of word vectors, which called context vectors, where each column represents a context word vector.

Formal training equation for Skip-Gram structure is given in the paper as following:

$$\frac{1}{T}\sum_{t=1}^{T}\sum_{-c \leq j \leq c, j \neq 0} \log p(w_{t+j}|w_t)$$

where T is the word size in training data and c is the context size of the target word at each side, in other words, half of the window size. p function in this equation refers to softmax function, which is given as:

$$p(w_O|w_I) = \frac{\exp(v'_{w_O}{}^\top v_{w_I})}{\sum_{w=1}^{W}\exp(v'_w{}^\top v_{w_I})}$$

where $v_w$ and $v'_w$ are the vectors from the first and second layer parameter matrices for word w and W is the vocabulary size.

In the second study (Mikolov et al., 2013b), it has stated that softmax calculation is expensive and to avoid that cost it has followed another method, negative sampling. In negative sampling, instead of calculating scores for each word in the vocabulary, random k negative sample are selected and the model is trained regarding only these negative samples to be minimized and positive samples to be maximized. It has also stated that a pre-processing for phrase detection and elimination of too frequent words.

### 3.2. fastText

Bojanowski et al. [14] proposed an improved version of CBOW/Skip-Gram models with "subword" information with the idea of using orthographical structure of words when training vectors, since morphological features cannot be preserved in plain CBOW/Skip-Gram models.

In this model, dictionary is made of subwords of words. A subword is simply a character n-gram, a part of a word where 3≤n≤6 is preferred in the study. First '<' and '>' characters are attached to beginning and end of words, then all subwords with all preferred n's are obtained and put it into a "bag". The word itself is also added to this bag of character n-grams. For instance, for the word "where" with 3≤n≤4, the bag should be as following:

[*<wh, whe, her, ere, re>, <whe, wher, here, ere>, <where>*]

Thanks to '<' and '>' characters, the word "her" and the subword "her" are represented differently.

In training, after obtaining the vocabulary of subwords, a vector is trained for each subword. A word vector is calculated by summing subword vectors. This way, even vectors of words that are never seen before can be calculated.

## 4. EMBEDDING MODELS

We used eight different embedding sets trained on the same corpus in our tests. Unless stated otherwise, all embeddings are trained by word2vec CBOW model with vector size of 300 and other options with default values.
A morphological disambiguator is used on the corpus to obtain lemma forms and suffixes of each word.

### 4.1. Surface Embeddings (Baseline)

The corpus is a plain text with surface (inflected) forms of the words. Vocabulary size of this model is 2,009,255.

### 4.2. Lemma Embeddings

Each word in the corpus has its inflection removed, in other words, is replaced with its lemma form. Example:

*Original sentence:* "Sokaklarda şarkı söyleyerek yürüdü ."
*English translation:* "She walked on streets by singing ."
*Lemma sentence:* "Sokak şarkı söyle yürü ."
*English translation:* "She walk street sing ."

When Lemma embeddings are tested in tasks, all surface forms of a word used the same Lemma embedding. This way we investigated any possible positive outcomes from removal of morphemes in Turkish texts.

Vocabulary size of this model is 1,098,219.

### 4.3. Lemma and Suffix Embeddings

Each word in the corpus is split as lemma and suffix, and suffixes are replaced with their id numbers in the suffix vocabulary prepared beforehand. Suffix numbers are prefixed with ">>" to not be confused with real numbers. Words without any suffix are also followed by a suffix number indicating they have no suffix. Example:

*Original sentence:* "Sokaklarda şarkı söyleyerek yürüdü ."
*Lemma and Suffix text:* "Sokak >>34 şarkı >>64 söyle >>11 yürü >>90 . >>41"

By its morphological nature, Turkish has a difficulty that prevents words that are close in semantics to occur in lemma forms together. One example can be words "car" ("araba") and "steering wheel" ("direksiyon"). While these two terms should be close in vector space, because they are used together with their inflected forms ("arabanın direksiyonu" / "steering wheel of car"), their lemma forms are not trained together, therefore their vectors don't share semantic similarities. With this method, we investigated if we can see improvements on vector semantics when we guarantee that each neighbor word in a context is trained with their lemma forms are exposed. Since this model halves the context window, the sliding windows is set to 9 instead of 5 in the training.

The model is tested in two methods. In the first one, only lemma vector is used for a word. Differently from Lemma embeddings, lemma vectors obtained from Lemma and Suffix model had suffixes in the context in training. In the second method, a word vector is determined by getting average of lemma and suffix vectors.

Vocabulary size of this model, which both contains lemma and suffix vectors, is 1,122,622

### 4.4. Derived Embeddings

In Derived embeddings, we derived suffix embeddings separately. For each suffix, we collected all words that include the suffix. Then we obtained the words' lemma forms and subtracted the lemma form vector from the surface form vector for each word. We got the average of these subtraction vectors and assigned it as the vector of the suffix.

$$vector(s) = \frac{1}{N} \sum_{i=1}^{N} vector(w_{si}) - vector(lemma(w_{si}))$$

where $w_s$ notates the words that have the suffix s.

When we need a vector of a word in tests, we separated it to lemma and suffix and added derived suffix vector to the word's lemma vector from Lemma embeddings.

$$vector(w) = vector\big(lemma(w)\big) + vector(suffix(w))$$

Vocabulary size for derived suffix vectors is 20,368.

### 4.5. SentencePiece Embeddings

SentencePiece embeddings are produced by preprocessing the corpus with SentencePiece[1] tokenizer. Differently from fastText embeddings, subword units are treated as independent tokens in the training, regardless of the word from which they are produced. In testing, words from the test sets are tokenized by SentencePiece and their vectors are averaged to get word vector.

$$vector(text) = \frac{1}{N} \sum_{i=1}^{N} vector(piece_i)$$

Vocabulary size for SentencePiece vectors is 31,993.

---

[1] https://github.com/google/sentencepiece

### 4.6. fastText Embeddings

fastText embeddings are produced by fastText model trained by using official python library. The corpus is a plain text with surface (inflected) forms of the words.
Vocabulary size for this model is 2,009,255.

## 5. EXPERIMENTS

### 5.1. Training Set

The word vectors are trained on a corpus consists of 950M words. The corpus is taken from the study of Güngör and Yıldız [8] which is publicly available[2].

### 5.2. Test Sets

Four tasks are selected for evaluating these vector sets: analogy task, text classification, sentimental analysis and language modeling.

For analogy task, we used semantic, syntactic and group analogy questions. Semantic and syntactic analogy questions contain two pairs of words with the same semantic or syntactic relations between them. Pairs are in the format as "house, houses, goal, goals" or "father, man, mother, woman" in Turkish. Test is guessing the last word given the first three. Testing is done with a formula of "vector(second word) - vector(first word) + vector(third word)" and expecting the vector of the last word to be the closest vector to this result vector, or to be in set of closest n vectors. For group questions, a group of words is given and the different one is expected to be predicted. One example can be "kilometer stere gram ounce quintal kiloton".

Data sets for these analogy tasks are obtained and merged from the studies of Güngör and Yıldız [8] and from Sen and Erdogan [7], all publicly available[3,4]. There are in total 45000 syntactic, 16000 semantic and 2000 group questions.

For text classification, we used 100 news from 13 different categories. For sentiment analysis, we used 2100 comments for 45 different books tagged with "positive", "negative" and "neutral". Both resources are obtained from Kemik NLP Research Group[5]. For language modeling, 1000 sentences are used for testing. For text classification and for sentiment analysis, one layer LSTM classifier is used. For language model, one layer RNN model is used. Inputs for these models are chosen as 20 words. For out-of-vocabulary words, a pre-determined word vector is used.

### 5.3. Results

#### 5.3.1. Analogy Task Results

Accuracy results for analogy task are shown in Table 1, Table 2 and Table 4.

---

[2]https://github.com/onurgu/linguistic-features-in-turkish-word-representations/releases
[3]http://myweb.sabanciuniv.edu/umutsen/research/
[4]https://github.com/onurgu/linguistic-features-in-turkish-word-representations/releases
[5]http://www.kemik.yildiz.edu.tr/?id=28

In syntactic test, we see different results based on the occurrence place. For the occurrence in the first prediction, we see that average vectors show the best result, followed by derived vectors. Both showed better results than surface vectors. When we consider appearance in the first X prediction, however, we see that new embedding models cannot compete with surface embeddings. In other words, when we need the target word to be the closest to the calculated word, average embeddings shows the best result but when we can afford the true word in the first 5 or 10 closest words, surface vectors provides better.

Table 1. Results for syntactic analogy task set.

| Embedding Model | Occurrence in the first n prediction | | | | | |
| --- | --- | --- | --- | --- | --- | --- |
| | 1 | 3 | 5 | 10 | 20 | 40 |
| Surface | 0.36 | **0.51** | **0.57** | **0.65** | **0.71** | **0.75** |
| Lemma & Suffix (average vector) | **0.40** | 0.43 | 0.44 | 0.46 | 0.47 | 0.50 |
| Derived Suffix | 0.39 | 0.43 | 0.44 | 0.46 | 0.49 | 0.51 |
| SentencePiece | 0.21 | 0.30 | 0.36 | 0.44 | 0.53 | 0.61 |
| fastText | 0.10 | 0.13 | 0.15 | 0.17 | 0.20 | 0.23 |

In semantic test, we see that embedding models which created by calculations fail very badly to keep semantic meanings. Surface embeddings shows the best result here, even though it is still lower than syntactic results.

Table 2. Results for semantic analogy task set.

| Embedding Model | Occurrence in the first n prediction | | | | | |
| --- | --- | --- | --- | --- | --- | --- |
| | 1 | 3 | 5 | 10 | 20 | 40 |
| Surface | **0.21** | **0.33** | **0.38** | **0.43** | **0.49** | **0.54** |
| Lemma | 0.10 | 0.14 | 0.16 | 0.18 | 0.20 | 0.22 |
| Lemma & Suffix (lemma only) | 0.07 | 0.09 | 0.11 | 0.13 | 0.14 | 0.16 |
| Lemma & Suffix (average vector) | 0 | 0.01 | 0.01 | 0.01 | 0.01 | 0.03 |
| Derived Suffix | 0 | 0 | 0 | 0 | 0.01 | 0.01 |
| SentencePiece | 0.03 | 0.05 | 0.06 | 0.08 | 0.11 | 0.15 |
| fastText | 0.01 | 0.02 | 0.02 | 0.04 | 0.06 | 0.09 |

To investigate the reason of the failure, we checked the result of a sample manually and compared the closest five words. The sample word set was "teyze, dayı, anne, baba" which are respectively "aunt, uncle, mother, father". In the task of predicting the forth word (baba), the results for the first five prediction of each model are shown in Table 3.

Table 3. First five predictions for the word set "teyze, dayı, anne, baba", separated by comma.

| Embedding Models | Semantic Predictions |
| --- | --- |
| Surface | baba, oğul, anne, anne-, oğulların |
| Lemma | korkman, münevver, vilâyetnâmeleri, hyperion'un, atakan |
| Lemma & Suffix (lemma only) | oğul, amca, baba, dede, yeğen |
| Lemma & Suffix (average vector) | amca, yeğen, geyik, büyükbaba, şaka |
| Derived Suffix | anneciğin, annenize, anneciğinin, annen, annenizde |
| SentencePiece | dayıyı, dayınızın, dayılarınızın, dayıyarak, dayıveya |
| fastText | dayız, dayın, dayını, dayısıydı, dayık |

It can be seen that lemma vectors of Lemma & Suffix embeddings and Surface vectors predicted family-related words, including the target word. Average vector predictions contains three relative nouns without the target word. Lemma vectors predicted completely irrelevant words.

Derived vectors always predicted inflected forms of the third word, which is a failure for this test. We can see that inflections of a word are gathered successfully but meaning relations could not be kept in vector differences, which resulted in pointing different inflections of the same word. These embedding models were trained on word parts divided by supervised morphological structures.

The vectors that are trained on unsupervised word parts, SentencePiece and fastText, similarly predicted the words that contains the second word. Differently from Derived vectors, the predictions were not always proper grammatical inflections, but also the words that just happened to contain the word by coincidence or typo.

In an overall look to the results of vector models built by partitioning, we can deduce that the reason of failure in semantic matches is mainly the impact of word derivations.

Table 4. Results for group analogy task set.

| **Embedding Models** | **Accuracy** |
|---|---|
| Surface | **0.34** |
| Lemma | 0.28 |
| Lemma & Suffix (lemma only) | **0.34** |
| Lemma & Suffix (average vector) | 0.24 |
| Derived Suffix | 0.29 |
| SentencePiece | 0.24 |
| fastText | 0.29 |

In group test, surface embeddings and Lemma vectors of Lemma and Suffix embeddings shows the same achievement, with a gap with the rest of the models.

It should be noted that when we use fastText vectors only with words those surface forms are in the vocabulary, it surpasses all models in syntactic and group tests, however when it is tested as it is intended (out-of-vocabulary word vectors are calculated by their subword units), the model's score lowered down.

### 5.3.2. Text Classification Results

Accuracy results for text classification are shown in Table 5.

Table 5. Results for text classification.

| **Embedding Models** | **Accuracy** |
|---|---|
| Surface | 0.90 |
| Lemma | 0.66 |
| Lemma & Suffix (lemma only) | 0.54 |
| Lemma & Suffix (average vector) | 0.52 |
| Derived Suffix | 0.76 |
| SentencePiece | **0.92** |
| fastText | **0.92** |

For text classification test, SentencePiece and fastText embeddings the best result, closely followed by Surface embeddings. It should be noted that SentencePiece model achieved this result with a small vocabulary compared to the others, which makes it the winner of this test.

### 5.3.3. Sentimental Analysis Results

Accuracy results for sentimental analysis are shown in Table 6.

Table 6. Results for sentimental analysis.

| Embedding Models | Accuracy |
|---|---|
| Surface | **0.73** |
| Lemma | 0.58 |
| Lemma & Suffix (lemma only) | 0.52 |
| Lemma & Suffix (average vector) | 0.53 |
| Derived Suffix | 0.60 |
| SentencePiece | **0.73** |
| fastText | 0.70 |

Differently from text classification, in sentimental analysis Surface and SentencePiece embeddings produce the best score, closely followed by fastText. Similar to the text classification, SentencePiece model achieves this with a very small vocabulary.

### 5.3.4. Language Modeling Results

Perplexity results for language modeling are shown in Table 7.

Derived suffix showed the lowest perplexity in language modelling task, very closely followed by SP embeddings with a negligible difference.

Table 7. Results for language modeling.

| Embedding Models | Perplexity |
|---|---|
| Surface | 13417 |
| Lemma | 12682 |
| Lemma & Suffix (lemma only) | 12606 |
| Lemma & Suffix (average vector) | 13179 |
| Derived Suffix | **12307** |
| SentencePiece | 12350 |
| fastText | 12856 |

## 6. CONCLUSIONS AND FUTURE WORK

Our study showed that manipulating words' surface forms can indeed improve the effectiveness of text representation a morphologically rich language. Training on words split into lemma and suffixes showed improvement to preserving syntactic features in word vectors and SentencePiece embeddings showed consistent top results on extrinsic task results compared to the other models. Even though all embeddings were trained by the same one-layer neural network, changing the input representation could show improvements. When context-free models are subject, because of their cheap training time, word2vec model trained on SentencePiece encodings gives the best outcome in extrinsic tasks.